# A Reliable Gravity Compensation Control Strategy for dVRK Robotic Arms With Nonlinear Disturbance Forces

Hongbin Lin, Chiu-Wai Vincent Hui, Yan Wang ⓘ, Anton Deguet ⓘ, Peter Kazanzides ⓘ, and K. W. Samuel Au ⓘ

*Abstract*—External disturbance forces caused by nonlinear springy electrical cables in the master tool manipulator (MTM) of the da Vinci Research Kit (dVRK) limits the usage of the existing gravity compensation methods. Significant motion drifts at the MTM tip are often observed when the MTM is located far from its identification trajectory, preventing the usage of these methods for the entire workspace reliably. In this letter, we propose a general and systematic framework to address the problems of the gravity compensation for the MTM of the dVRK. Particularly, high-order polynomial models were used to capture the highly nonlinear disturbance forces and integrated with the multi-step least square estimation framework. This method allows us to identify the parameters of both the gravitational and disturbance forces for each link sequentially, preventing residual error passing among the links of the MTM with uneven mass distribution. A corresponding gravity compensation controller was developed to compensate the gravitational and disturbance forces. The method was validated with extensive experiments in the majority of the manipulator's workspace, showing significant performance enhancements over existing methods. Finally, a deliverable software package in MATLAB and C++ was integrated with dVRK and published in the dVRK community for open-source research and development.

## I. INTRODUCTION

DURING a tele-operated surgery, surgeons usually operate the robots at a relatively low speed for a prolonged period. It is exceptionally burdensome for users to lift the weight of the master devices of a surgical robot, which can lead to muscle fatigue and lower surgery performance. As such, reliable gravity compensation for the master devices of tele-surgical systems for the entire workspace is important for both ergonomic and safety reasons.

Due to its large workspace, full actuation, and ease-of-use, the Master Tool Manipulator (MTM) of the da Vinci Research Kit (dVRK) [1] has become a popular option for user input or haptic devices in the surgical robotic research community [2], [3]. Although the dynamic parameter identification and gravity compensation for general serial manipulators have been studied extensively [4], [5], their application to the MTM of the dVRK poses different challenges, including the modeling of closed-loop kinematic chains and particularly the external disturbances created by the nonlinear springs and electrical cables through multiple joints [6], [7]. For this reason, the existing software packages for the dynamic model identification of general serial manipulators, such as SymPybotics [8] and FloBaRoID [9], cannot be applied to the MTM directly.

Researchers have proposed various methods to address this problem [6], [10]–[12]. In [6], the dynamic parameters were calculated based on simple geometric models using a Computer Aided Design (CAD) method accompanied with extensively manual tuning. [10] applied a standard Single-step Least Square Estimation (SLSE) approach to systematically identify the dynamic parameters. [11] applied the methods proposed by [13] to identify the dynamic parameters of the dVRK arms with physical consistency, using Semi-definite Programming and Linear Matrix Inequality techniques. However, all of the aforementioned work failed to model the significant external disturbance forces caused by electric cables.

Although significant strides have been made, these methods still have not been widely used in the dVRK community during the daily operation. The highly nonlinear, configuration-dependent and direction-dependent external disturbance forces caused by the combination of the electrical cables and spring on the MTM of the dVRK remain the main hurdle for the wide adoption of these methods (See Fig. 2). Experimental studies in [7], [10] showed that, with these methods, significant motion drifts at the MTM tip were often observed when the MTM was located far from its identification trajectory, preventing the reliable usage of these methods for the entire workspace. [11] proposed to compensate for the elastic force of the external electrical cables with a simple linear spring model. However, the significant unmodeled cable nonlinearities limit the effectiveness of the proposed method (Fig. 3).

Manuscript received February 24, 2019; accepted June 19, 2019. Date of publication July 10, 2019; date of current version August 2, 2019. This letter was recommended for publication by Associate Editor E. De Momi and Editor P. Valdastri upon evaluation of the reviewers' comments. This work was supported in part by the CUHK Chow Yuk Ho Technology Centre of Innovative Medicine, in part by SHIAE (BME-p1-17), and in part by the Natural Science Foundation of China under Grant U1613202. *(Corresponding author: K. W. Samuel Au.)*

H. Lin, C.-W. Vincent Hui, Y. Wang, and K. W. S. Au are with the Department of Mechanical and Automation Engineering, Chinese University of Hong Kong, Hong Kong (e-mail: hongbinlin@cuhk.edu.hk; vincenthk007@gmail.com; wangyanhiter@gmail.com; samuelau@cuhk.edu.hk).

A. Deguet and P. Kazanzides are with the Department of Computer Science, Johns Hopkins University, Baltimore, MD 21218 USA (e-mail: anton.deguet@jhu.edu; pkaz@jhu.edu).

This letter has supplementary downloadable material available at http://ieeexplore.ieee.org, provided by the authors. The video is designed for a better demonstration on the motivation, proposed experiments and the effectiveness of our method. Contact samuelau@cuhk.edu.hk for further questions about this letter.

Digital Object Identifier 10.1109/LRA.2019.2927953

[12] was the first to use a high order polynomial function to approximate the nonlinear disturbance force created by electrical cables on Joint 4 of the MTM. However, this primitive approach requires to set the axis of Joint 4 parallel to the gravity so as to decouple the polynomial parameter estimation from the gravitational effect. This requirement poses a great challenge to extending this method to all links since some joint axes (e.g. the axis of Joint 3) cannot be parallel to gravity due to the mechanical constraints of the MTM. Therefore, it is required to extend the modeling of disturbance forces with more complicated models and integrate it into the parameter identification procedure for all links.

To the best of our knowledge, there has not been a reliable solution for the gravity compensation of the MTM for a large range of workspace. More importantly, there is an urgent need to create a reproducible and easy-to-use software package for the community. Hence, we present such a gravity compensation solution for the dVRK MTM to address the technical issues of the unmodeled nonlinear disturbance forces and joint drifts in existing methods. Overall, the key contributions are:

1) Proposal of a systematic framework for parameter estimation and gravity compensation controller (GCC), which uses polynomial functions to model the non-linear disturbance behaviors, and effectively integrates with the Multi-step Least Square Estimation (MLSE) framework (Section III) for a general multi-DOF serial manipulators with nonlinear, configuration-dependent and direction-dependent disturbance forces across joints. This method decouples the multi-DOF estimation into a sequential joint-to-joint estimation for both the gravitational and disturbance forces.
2) Proposal of a two-joint moving data collection strategy, where two joints move simultaneously to lower down the condition number and estimation error as compared to a one-joint motion approach.
3) Demonstration of the effectiveness of the GCC compared with the recent state-of-art methods. Through extensive experiments, the proposed method was verified to be effective in providing precise gravity compensation in the majority of the workspace. A dVRK gravity compensation software package was published to the dVRK community [14].

## II. MODELING OF THE MASTER TOOL MANIPULATOR

In this section, we first derive the kinematics of the MTM to obtain the coordinate of the center of mass (COM) of each link. Then, we introduce the dynamic modeling of the MTM, particularly, focusing on the modeling of the gravitational force and external disturbance forces.

### A. Kinematic Modeling

The MTM is a manipulator with seven motorized revolute joints and one passive pinching joint. The pinching joint is neglected in our modeling since it is not motorized. The last actuated joint is also not considered in our kinematic modeling since the gravitational force has no effect on this joint due to the symmetrical structure of its link about its joint axis, and the external disturbance from cables is minimal for this joint. Therefore, the system is equivalent to a 6-DOF serial manipulator with the mass of Link 6 lumped with the last 2 links. Fig. 1 shows the coordinate frame definition of the left MTM according to Denavit-Hartenberg (DH) convention [15]. Two serial chains are employed to model the MTM and its parallel mechanism. The detailed DH parameters for the MTM and its parallel mechanism can be found in our dVRK gravity compensation webpage on GitHub [14].

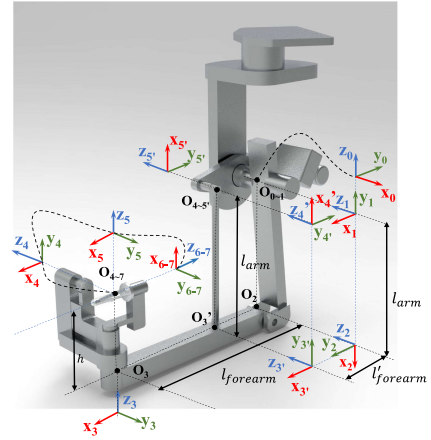

Fig. 1. DH frames of the MTM.

The coordinate of the COM of a serial manipulator link with respect to the base frame can be written in homogeneous representation and calculated recursively by

$$\boldsymbol{p}_{c_i}(q_1,..,q_i) = \left(\prod_{k=1}^{i} \boldsymbol{A}_k^{k-1}(q_k)\right) \boldsymbol{r}_i = [x_{c_i}\; y_{c_i}\; z_{c_i}\; 1]^T, \quad (1)$$

where $\boldsymbol{A}_{i-1}^i(q_i)$ is the homogeneous transformation matrix from Frame $i-1$ to Frame $i$ based on [15], and $\boldsymbol{r}_i$ is the COM of Link $i$ represented in Frame $i$.

### B. Dynamic Modeling

In the application of gravity compensation, the robot dynamics can be simplified as a static model, which mainly contains the gravitational force and external disturbance forces.

*1) Modeling of the Gravitational Force:* Using Euler-Lagrange approach [15], the manipulator dynamics can be formulated as

$$\boldsymbol{\tau} = \frac{d}{dt}\frac{\partial L}{\partial \dot{\boldsymbol{q}}} - \frac{\partial L}{\partial \boldsymbol{q}}, \quad (2)$$

where $L, \boldsymbol{q}, \boldsymbol{\tau}$ are the Lagrangian, joint angles, and joint torques, respectively. Here, the Lagrangian $L$ is defined as $L = K - P$, where $K$ and $P$ represent the total kinetic and potential energy of the robot, respectively. In the static condition, the kinetic energy of the robot is zero ($K = 0$), and the Lagrangian $L$ can then be written as

$$L(\boldsymbol{q}) = -P = -\sum_{i=1}^{n} m_{c_i} z_{c_i} g, \quad (3)$$

where $g$ is the gravitational acceleration constant, $z_{c_i}$ is calculated using (1), $m_{c_i}$ is the mass of Link $i$, and $n$ is the number of links of a serial chain. Using $K = 0$ and (2), the joint torque from the gravitational force $\boldsymbol{\tau}_g$ becomes

$$\boldsymbol{\tau}_g = -\frac{\partial L}{\partial \boldsymbol{q}}. \quad (4)$$

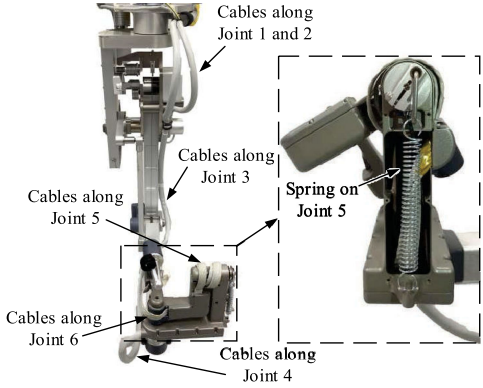

Fig. 2. Sources of the nonlinear external disturbance forces. The electrical cables pass along each MTM joint and create disturbances force on them. The internal spring of Joint 5 combined with the geometry becomes another source of nonlinear disturbance force to its joint.

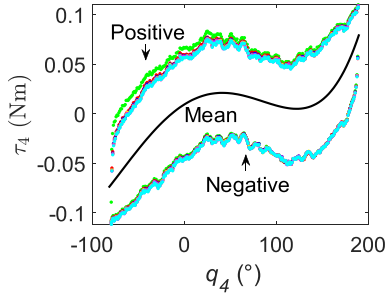

Fig. 3. Torque vs position on joint 4.

*2) Modeling of the External Disturbance Force:* Other than the gravitational force, there exists significant static joint torque that needs to be modeled and compensated precisely to improve the GCC performance [11], [12]. This static torque is mainly created by the configuration-dependent and direction-dependent force from the electric cables adhered to each joint (Fig. 2). The internal spring of Joint 5 combined with the geometry becomes another source of nonlinear disturbance forces to its joint. Fig. 3 shows the typical experimental position-torque curve of Joint 4. During the data collection, the axis of Joint 4 was set parallel to the gravitational force, and the robot was stationary to minimize the effect of viscous friction and inertia. As can be seen in Fig. 3, the disturbance force is highly nonlinear and varies with the joint movement position and direction.

Using these insights, we propose a general model representation of the external disturbance for a single joint. In this model, $\tau_{ec}$ and $\tau_{ed}$ represent the joint torques from the configuration-dependent and direction-dependent effects of disturbances, respectively. We use these joint torques to represent the external disturbance forces, $\tau_{ext}^+$ and $\tau_{ext}^-$, in the positive and negative directions as

$$\tau_{ext}^+ = \tau_{ec} + \tau_{ed}, \quad \tau_{ext}^- = \tau_{ec} - \tau_{ed} \quad (5)$$

Instead of the physics-based modeling approach, we use polynomial functions to approximate the highly nonlinear behaviour of the external disturbance forces. The joint torque from the external disturbance forces on one joint is thus represented as a product of a $k^{\text{th}}$ order polynomial function of the joint angle and the corresponding joint movement direction selection term as

$$\begin{aligned}\tau_{ext} &= \tau_{ext}^+ u(\Delta q) + \tau_{ext}^-(1 - u(\Delta q)) \\ &= \boldsymbol{\phi}^T \boldsymbol{a}^+ u(\Delta q) + \boldsymbol{\phi}^T \boldsymbol{a}^- (1 - u(\Delta q)) \\ &= \boldsymbol{\phi}^T(\boldsymbol{a}^+ u(\Delta q) + \boldsymbol{a}^-(1 - u(\Delta q)))\end{aligned} \quad (6)$$

where $\boldsymbol{\phi} = \begin{bmatrix} 1 & q^1 & \ldots & q^k \end{bmatrix}^T$ is the vector related to joint angle $q$, $\Delta q$ is the difference between the current joint angle and the joint angle in the last control iteration, and $\boldsymbol{a}^+$ and $\boldsymbol{a}^-$ are the polynomial coefficient vectors in the positive and negative directions, respectively. $u(\bullet)$ is the step function used to determine the direction of the joint movement.

In this general representation (6), two separate sets of $k^{\text{th}}$ order polynomials are used to capture the direction-dependent behavior of the disturbance forces. The choice of the order of the polynomial functions for each joint is determined based on empirical study. More details can be found in Section VI.

### III. PARAMETER ESTIMATION

In this section, we first present the integration of the aforementioned models with the standard SLSE approach and then extend it towards the MLSE method.

#### A. SLSE With Gravitational and Nonlinear Disturbance Forces

The gravitational torque of the MTM in (4) can be represented as a linear combination of its dynamic parameters [4] such that.

$$\boldsymbol{\tau}_g = {}^g\boldsymbol{Y}(\boldsymbol{q}){}^g\boldsymbol{\beta} \quad (7)$$

where ${}^g\boldsymbol{Y} \in \mathbb{R}^{n \times b}$ and ${}^g\boldsymbol{\beta} \in \mathbb{R}^b$ are the regressor and the lumped dynamic parameter vector for the gravitational torque, with $n$ and $b$ being the numbers of joints and lumped dynamic parameters, respectively. This formulation can be obtained numerically through QR decomposition [16]. The regressor, ${}^g\boldsymbol{Y}$, is latter used for the estimation of ${}^g\boldsymbol{\beta}$. The explicit form of ${}^g\boldsymbol{Y}$ and ${}^g\boldsymbol{\beta}$ are not presented in the letter due to the page limit and can be found in the Github web page [14].

Since we consider different polynomial coefficients in different joint movement directions, the parameters for the external disturbance forces ${}^{ext}\boldsymbol{\beta}$ are represented as

$${}^{ext}\boldsymbol{\beta} = \begin{bmatrix} ({}^{ext}\boldsymbol{\beta}^+)^T & ({}^{ext}\boldsymbol{\beta}^-)^T \end{bmatrix}^T, \quad (8)$$

where ${}^{ext}\boldsymbol{\beta}^+$ and ${}^{ext}\boldsymbol{\beta}^-$ are the parameter vectors for the positive and negative joint movement directions, respectively, and represented as

$$\begin{aligned}{}^{ext}\boldsymbol{\beta}^+ &= \begin{bmatrix} (\boldsymbol{a}_1^+)^T & (\boldsymbol{a}_2^+)^T & \ldots & (\boldsymbol{a}_n^+)^T \end{bmatrix}^T \\ {}^{ext}\boldsymbol{\beta}^- &= \begin{bmatrix} (\boldsymbol{a}_1^-)^T & (\boldsymbol{a}_2^-)^T & \ldots & (\boldsymbol{a}_n^-)^T \end{bmatrix}^T\end{aligned} \quad (9)$$

where $\boldsymbol{a}_i^+$ and $\boldsymbol{a}_i^-$ are the parameters for the external disturbance force for Joint $i$ in positive and negative movement directions, respectively, and $n$ is the number of DOFs of the manipulator.

As either $\boldsymbol{a}_i^+$ or $\boldsymbol{a}_i^-$ works at a certain time, we define a diagonal activation matrix $\boldsymbol{U}(\boldsymbol{\Delta q})$ to select the right polynomial coefficients to use, based on the joint angle differences between two successive control iterations $\Delta q$, as

$$U(\Delta q) = \mathrm{diag}\{\underbrace{u(\Delta q_1),\ldots,u(\Delta q_1)}_{k+1},\ldots,\underbrace{u(\Delta q_n),\ldots,u(\Delta q_n)}_{k+1}\} \quad (10)$$

where $k$ is the order of the polynomial functions.

The joint torque caused by the external disturbance forces $\tau_{ext}$ can then be expressed in a linear form of the external disturbance parameters $^{ext}\beta$ as

$$\tau_{ext} = {}^{ext}Y(q,\Delta q){}^{ext}\beta$$
$$= [\Phi(q)U(\Delta q) \quad \Phi(q)(1 - U(\Delta q))]\begin{bmatrix}{}^{ext}\beta^+ \\ {}^{ext}\beta^-\end{bmatrix}, \quad (11)$$

where $\Phi$ is the regressor matrix for the external disturbance forces

$$\Phi(q) = \begin{bmatrix} \phi_1^T & 0 & \cdots & 0 \\ 0 & \phi_2^T & \cdots & 0 \\ \vdots & \vdots & \ddots & \vdots \\ 0 & 0 & \cdots & \phi_n^T \end{bmatrix}, \quad (12)$$

where $\phi_i$ is the polynomial terms of Joint $i$ in (6).

Using (7) and (11), we can obtain the dynamic equation considering both the gravitational force and external disturbance force as

$$\tau = \tau_g + \tau_{ext} = Y(q,\Delta q)\beta = [{}^gY \quad {}^{ext}Y]\begin{bmatrix}{}^g\beta \\ {}^{ext}\beta\end{bmatrix}, \quad (13)$$

where $\beta \in \mathbb{R}^{m\times 1}$ and $Y \in \mathbb{R}^{n\times m}$ are the augmented dynamic parameter vector and dynamic regressor matrix, respectively, and $m$ is the number of dynamic parameters.

With the collected data, the dynamic regressor matrices and joint torque vectors of the training data can be stacked into an augmented form as

$$W = \begin{bmatrix} Y(q^1,\dot{q}^1) \\ Y(q^2,\dot{q}^2) \\ \vdots \\ Y(q^p,\dot{q}^p) \end{bmatrix}, \quad \omega = \begin{bmatrix} \tau^1 \\ \tau^2 \\ \vdots \\ \tau^p \end{bmatrix}, \quad (14)$$

where $W$ and $\omega$ are data regressor matrix and data torque vector, respectively. $q^i$ and $\tau^i$ are the joint angle and torque vector at the $i^{th}$ sampling point of training data. $p$ is the total number of training data.

Least Square Estimation is then applied to estimate the dynamic parameter vector $\hat{\beta}$ [15].

$$\hat{\beta} = (W^TW)^{-1}W^T\omega = W^\dagger\omega, \quad (15)$$

where $W^\dagger$ is the left pseudo-inverse of $W$.

### B. MLSE With Gravitational and Nonlinear Disturbance Forces

In SLSE, the parameter estimation for all joints is coupled together to minimize the residual error of joint torques over all joints, hence, the estimated dynamic parameters of different links can affect each other. [5] proposed an iterative approach from a mechanics perspective to estimate the gravity compensation parameters by fitting the joint torque of each joint with a sinusoidal function of joint angle. Although the principle behind [5] and our approach is the same, we derive it from a different perspective. Moreover, we integrate the external disturbance force model with our proposed MLSE.

The gravitational force of Joint $i$ only depends on the dynamic parameters of the current link and its child links, i.e., Link $i$ to 6, and is independent of its parent links, i.e., Link 1 to $i-1$. Therefore, the dynamic regressor matrix for the gravitational force ${}^gY$, as shown in (16), appears in an upper echelon manner from the $2^{nd}$ row to the $6^{th}$ row. Meanwhile, the external disturbance force of Joint $i$ only depends on the parameters of Joint $i$, based on (12). Using these properties, we can divide the estimation process into multiple iterative steps from the distal joints to the proximal joints sequentially.

$${}^gY = \begin{bmatrix} 0 & 0 & 0 & 0 & 0 & 0 & 0 & 0 & 0 & 0 \\ * & * & \# & \# & \# & \# & \# & \# & \# & \# \\ 0 & 0 & * & * & \# & \# & \# & \# & \# & \# \\ 0 & 0 & 0 & 0 & * & * & \# & \# & \# & \# \\ 0 & 0 & 0 & 0 & 0 & 0 & * & * & \# & \# \\ 0 & 0 & 0 & 0 & 0 & 0 & 0 & 0 & * & * \end{bmatrix} \quad (16)$$

where $*$ and $\#$ correspond to the parameters to be estimated in current step and estimated in previous steps. The elements of the $1^{st}$ rows in $Y_g$ are zero because the axis of Joint 1 is along the direction of gravity.

To implement MLSE, we first define several terms. ${}^g\hat{\beta}_i$ and ${}^g\bar{\beta}_i$ are the dynamic parameters in ${}^g\beta$ to estimate for Joint $i$ and estimated prior to Joint $i$, which are corresponding to the terms of $*$ and $\#$ in (16), respectively. ${}^g\hat{Y}_i$ and ${}^g\bar{Y}_i$ are the terms in the $i^{th}$ row of ${}^gY$, consisting of the terms of $*$ and $\#$ in (16), respectively.

Considering both gravitational and external disturbance terms, we can write the parameters to estimate for Joint $i$, $\hat{\beta}_i$, and estimated prior to Joint $i$, $\bar{\beta}_i$, as

$$\hat{\beta}_i = \left[({}^g\hat{\beta}_i)^T \quad ({}^{ext}\hat{\beta}_i)^T\right]^T, \quad \bar{\beta}_i = {}^g\bar{\beta}_i, \quad (17)$$

where

$${}^{ext}\hat{\beta}_i = \left[(a_i^+)^T \quad (a_i^-)^T\right]^T. \quad (18)$$

The regressor matrices corresponding to $\hat{\beta}_i$ and $\bar{\beta}_i$ can be represented as

$$\hat{Y}_i = \left[{}^g\hat{Y}_i \quad {}^{ext}\hat{Y}_i\right], \quad \bar{Y}_i = {}^g\bar{Y}_i, \quad (19)$$

where

$${}^{ext}\hat{Y}_i = \left[\phi_i^T U(\Delta q_i) \quad \phi_i^T(1 - U(\Delta q_i))\right], \quad (20)$$

where $U(\Delta q_i) = \mathrm{diag}\{u(\Delta q_i),\ldots,u(\Delta q_i)\}$ is the $k+1$ dimensional diagonal activation matrix for Joint $i$. Based on (17) and (19), the $i^{th}$ row of (13) can be rewritten as

$$\tau_i = Y_i(q,\Delta q)\beta_i = [\hat{Y}_i \quad \bar{Y}_i]\begin{bmatrix}\hat{\beta}_i \\ \bar{\beta}_i\end{bmatrix} = \hat{Y}_i\hat{\beta}_i + \bar{Y}_i\bar{\beta}_i. \quad (21)$$

We also rewrite (21) as

$$\hat{Y}_i\hat{\beta}_i = \tau_i - \bar{Y}_i\bar{\beta}_i. \quad (22)$$

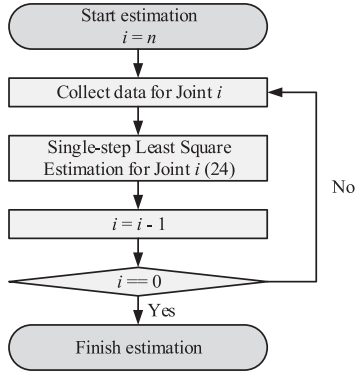

Fig. 4. Procedure of multi-step least square estimation.

TABLE I
JOINT RANGES FOR THE ESTIMATED AND AUXILIARY JOINTS OF THE MTM

| joint number | range (°) | auxiliary joint number | range (°) |
| --- | --- | --- | --- |
| 6 | [-40, 40] | 5 | [-90, 90] |
| 5 | [-85, 175] | 3 | [-10, 20] |
| 4 | [-190, 80] | 3 | [-10, 20] |
| 3 | [-34, 34] | 2 | [-14, 40] |
| 2 | [-14, 40] | - | - |
| 1 | [-7, 40] | - | - |

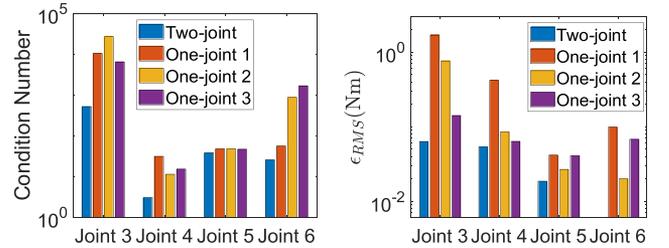

(a) Condition number of the regressor matrices with the sampling data from the one-joint and two joint approaches.

(b) RMS absolute error of the measured and predicted joint torque with the sampling data from the one-joint and two-joint approaches.

Fig. 5. Comparison of one-joint and two-joint data collection approaches. For the one-joint approach, three different auxiliary joint positions. ($a_1$, $a_2$, and $a_3$) were used, corresponding to the lower limit, midpoint, upper limit of the joint range shown in Table I. Particularly, when applying the two-joint data collection approach to Joint 6, the RMS absolute error was too small and beyond the current resolution of (b).

Thus, the regressor matrices for the parameters to estimate for Joint $i$ and estimated prior to Joint $i$, $\hat{\bm{W}}_i$ and $\bar{\bm{W}}_i$, and the data torque vector $\bm{\omega}_i$ of Joint $i$ can be rewritten as

$$\hat{\bm{W}}_i = \begin{bmatrix} \hat{\bm{Y}}_i(\bm{q}^1, \bm{\Delta q}_i^1) \\ \hat{\bm{Y}}_i(\bm{q}^2, \bm{\Delta q}_i^2) \\ \vdots \\ \hat{\bm{Y}}_i(\bm{q}^p, \bm{\Delta q}_i^p) \end{bmatrix}, \bar{\bm{W}}_i = \begin{bmatrix} \bar{\bm{Y}}_i(\bm{q}^1) \\ \bar{\bm{Y}}_i(\bm{q}^2) \\ \vdots \\ \bar{\bm{Y}}_i(\bm{q}^p) \end{bmatrix}, \bm{\omega}_i = \begin{bmatrix} \tau_i^1 \\ \tau_i^2 \\ \vdots \\ \tau_i^p \end{bmatrix} \quad (23)$$

where $\bm{q}^j$ is the joint angle of the robot, and $\tau_i^j$ is the measured torque of Joint $i$ at the $j^{\text{th}}$ sampling point of the training data.

Therefore, the regression problem for the parameters of Joint $i$ can be described as

$$\hat{\bm{\beta}}_i = \begin{cases} \hat{\bm{W}}_i^\dagger \bm{\omega}_i, & i = n \\ \hat{\bm{W}}_i^\dagger (\bm{\omega}_i - \sum_{k=i+1}^n \bar{\bm{W}}_k \bar{\bm{\beta}}_k), & i \neq n \end{cases} \quad (24)$$

The procedure of MLSE is illustrated in Fig. 4. Since we estimate the parameters for one link/joint in each step of MLSE, we can immediately conduct the validation of the estimated results through experiments (See the drift test in Section VI-B) for that link/joint before moving to the next MLSE step. With the integrated validation process, we can guarantee the accuracy of the estimation for each joint/link, preventing any unwanted estimated error accumulation spreading across joints as in SLSE method [7], [10]. The software troubleshooting and debugging also become easier to manage from an individual joint/link perspective.

## IV. DATA COLLECTION STRATEGY

In data collection, joint position and torque of the MTM are recorded at different combinations of joint positions. Generally, training samples should contain all the combinations of joint positions within the reachable joint space. However, data amount will increase exponentially in the scale of $O(N^n)$ if all possible joint combinations are explored for a manipulator with high DOFs, where $N$ is the number of sampled data for each joint.

In theory, moving Joint $i$ is sufficient for parameter estimation of Joint $i$ given that the data regressor matrix $\hat{\bm{W}}_i$ is not singular. However, due to the strong approximation capacity of polynomial functions, the torque data due to gravity force may be overfitted to the polynomial functions. Moving two non-parallel joints for data collection exposes the robots to more configurations and helps to reduce overfitting issues. Therefore, when we collect data to estimate the parameters of Joint $i$, the joint positions of two joints vary, including Joint $i$ and one of its parent joints, which we call the auxiliary joint of Joint $i$.

Table I shows the joint ranges of the estimated joints and their corresponding auxiliary joints of the MTM. As Joint 1 of the MTM is the first joint and its joint axis is along the direction of the gravitational force, the gravitational force has no effect on its joint torque. Moreover, Joint 1, the only parent joint of Joint 2, does not change the angle between the axis of Joint 2 and the gravitational force. Therefore, there are no auxiliary joints for the first two joints. By using this data collection strategy, we can shrink the scale of sampled data to $O(n \times N^2)$.

An experimental study was conducted to evaluate the effectiveness of the two data collection strategies: one-joint and two-joint moving data strategies. The condition number of the regressor matrices $\bm{W}_i$ for Joint $i$ formed by the two strategies were calculated and used as an index to indicate the extent of the excitation through each strategy [11], [13]. During the experiment, for the two-joint strategy, 400 data points were collected uniformly in the range of Joint $i$ and its auxiliary joint. In one-joint strategy, the same amount of data points were collected uniformly with only Joint $i$ changing in its joint range and the auxiliary joint set to either the lower limit ($a_1$), midpoint ($a_2$), and upper limit ($a_3$) of its range (Table I). At the estimation step for Joint $i$, the data was collected in two joint motion directions separately to estimate the parameters for the external disturbance force $^{ext}\bm{\beta}_i^+$ and $^{ext}\bm{\beta}_i^-$.

As shown in Fig. 5a, the condition numbers of the regressor matrices of the two-joint strategy are lower than those of the one-joint strategy for all joints. The corresponding Root Mean Square (RMS) absolute error results obtained through the two-joint strategy are also lower than those obtained through the one-joint approach (see Fig. 5b).

## V. Gravity Compensation Controller (GCC)

A GCC is implemented to compensate for the gravitational and external disturbance forces as

$$\boldsymbol{\tau} = \hat{\boldsymbol{\tau}}_g + \hat{\boldsymbol{\tau}}_{ext}, \quad (25)$$

where $\hat{\boldsymbol{\tau}}_g$ and $\hat{\boldsymbol{\tau}}_{ext}$ are the estimated joint torques from the gravitational and external disturbance forces, respectively.

The estimated joint torque from the gravitational force can be calculated as

$$\hat{\boldsymbol{\tau}}_g = {}^g\boldsymbol{Y}(\boldsymbol{q}){}^g\hat{\boldsymbol{\beta}}. \quad (26)$$

Based on our models in Section II, the external disturbance force consists of the configuration-dependent torque $\boldsymbol{\tau}_{ec}$ and direction-dependent torque $\boldsymbol{\tau}_{ed}$. The estimated configuration-dependent joint torque $\hat{\boldsymbol{\tau}}_{ec}$ can be calculated as the mean of the two polynomial functions in the positive and negative directions,

$$\hat{\boldsymbol{\tau}}_{ec} = (\hat{\boldsymbol{\tau}}_{ext}^+ + \hat{\boldsymbol{\tau}}_{ext}^-)/2 = \boldsymbol{\Phi}(\boldsymbol{q})({}^{ext}\hat{\boldsymbol{\beta}}^+ + {}^{ext}\hat{\boldsymbol{\beta}}^-)/2. \quad (27)$$

The estimated direction-dependent joint torque $\hat{\boldsymbol{\tau}}_{ed}$ can be calculated as half of the difference of the two polynomial functions in the positive and negative directions,

$$\hat{\boldsymbol{\tau}}_{ed} = (\hat{\boldsymbol{\tau}}_{ext}^+ - \hat{\boldsymbol{\tau}}_{ext}^-)/2 = \boldsymbol{\Phi}(\boldsymbol{q})({}^{ext}\hat{\boldsymbol{\beta}}^+ - {}^{ext}\hat{\boldsymbol{\beta}}^-)/2. \quad (28)$$

In our implementation, we set a dead band $\Delta q_{db_i}$ for Joint $i$ to compensate its direction-dependent force to remove the effect of the noise of the measured joint angle when $\Delta q$ is small. When the absolute value of $|\Delta q_i|$ is larger than the saturated value $\Delta q_{s_i}$, we only compensate a certain ratio $\alpha$ of all the estimated direction-dependent torque $\hat{\tau}_{ed_i}$ to avoid instability of the system. Between $\Delta q_{db_i}$ and $\Delta q_{s_i}$, a linear interpolation is used to ensure the continuity of the joint torque. Finally, the ratio of the joint torque to compensate the direction-dependent force of Joint $i$, $\xi_i$, is calculated as

$$\xi_i = \begin{cases} 0 & |\Delta q_i| \leq \Delta q_{db_i} \\ \frac{|\Delta q_i| - \Delta q_{db_i}}{\Delta q_{s_i} - \Delta q_{db_i}} \operatorname{sgn}(\Delta q_i)\alpha & \Delta q_{db_i} \leq |\Delta q_i| \leq \Delta q_{s_i} . \\ \operatorname{sgn}(\Delta q_i)\alpha & \Delta q_{s_i} \leq |\Delta q_i| \end{cases} \quad (29)$$

With $\xi_i$ for all the joints, we define a diagonal matrix $\boldsymbol{\xi}$ as the direction-dependent force compensation ratio matrix.

$$\boldsymbol{\xi} = \operatorname{diag}\{\xi_1, \xi_2, \ldots, \xi_n\}. \quad (30)$$

Therefore, the total torque to compensate is represented as

$$\boldsymbol{\tau}_c = \hat{\boldsymbol{\tau}}_g + \hat{\boldsymbol{\tau}}_{ec} + \boldsymbol{\xi}\hat{\boldsymbol{\tau}}_{ed}. \quad (31)$$

Fig. 6 shows the overview of our GCC. Joint positions are received from the MTM in real time. Matrices ${}^g\boldsymbol{Y}(\boldsymbol{q})$ and $\boldsymbol{\Phi}(\boldsymbol{q})$ are calculated using current joint position $\boldsymbol{q}$, and $\boldsymbol{\xi}$ is calculated using $\Delta\boldsymbol{q}$. The GCC integrated with the dVRK software in C++ operates at 2 kHz.

## VI. Experiments and Results

Based on the dVRK software system [1], we developed a software package written in MATLAB and C++ for the gravity compensation of the MTM, including data collection, parameter identification, and a GCC [14]. In this section, we will present the results of our parameter estimation and gravity compensation validation.

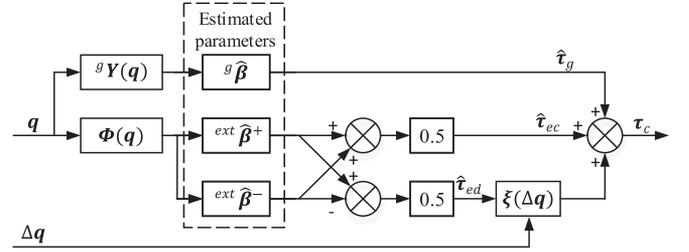

Fig. 6. Gravity compensation controller for the MTM.

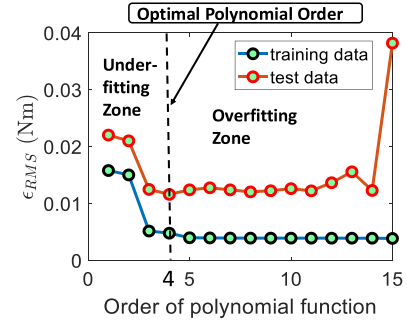

Fig. 7. RMS Absolute Error between the measured and predicted joint torque of Joint 5 modeled with different polynomial function orders.

### A. Parameter Estimation Validation

We applied our data collection strategy and estimation method to estimate the parameters for each joint. During the process, we collected data at 600 different joint configurations for each joint. RMS relative error between the measured and predicted joint torques of Joint $i$ was used to evaluate the estimation performance, which is defined as

$$\epsilon_{RMS_i}\% = ||\boldsymbol{W}_i\hat{\boldsymbol{\beta}} - \boldsymbol{\omega}_i||/||\boldsymbol{\omega}_i|| \cdot 100\%, \quad (32)$$

One key step is to choose an appropriate order for the polynomial functions for each joint. We approached this question from an empirical perspective. After data collection, for each joint, we ran our estimation algorithm multiple times with an increasing polynomial function order and plotted the RMS absolute error against the polynomial order to determine the empirical optimal order. Fig. 7 shows a typical curve of RMS absolute error against different polynomial function orders. As can be seen, as the order increases, the training error decreases all the time, while the test error decreases first and then starts to increase due to overfitting when the order is larger than a certain value.

After performing similar studies for all the joints of the MTM, we decided to use $4^{\text{th}}$ order polynomial functions to model the external disturbance for all the joints, except Joint 2 for which $1^{\text{st}}$ order is adequate. With the selected polynomial function order, we finalized the parameters of the MTM and detailed parameter values can be found in [14].

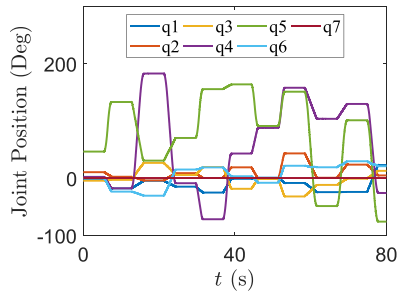

Fig. 8. Joint trajectory for the MTM in the trajectory test.

TABLE II
RMS RELATIVE ERROR BETWEEN THE MEASURED AND PREDICTED JOINT TORQUES IN THE TRAJECTORY TEST

| method | Joint 1 | Joint 2 | Joint 3 | Joint 4 | Joint 5 | Joint 6 |
|---|---|---|---|---|---|---|
| CAD (%) | 100 | 18.20 | 20.78 | 70.09 | 107.46 | 63.10 |
| Fontanelli *et al.* (%) | 67.92 | 17.62 | 21.66 | 47.05 | 88.92 | 32.06 |
| Our method (%) | **71.62** | **12.58** | **7.92** | **34.55** | **25.22** | **31.94** |

TABLE III
MAXIMUM ABSOLUTE ERROR BETWEEN THE MEASURED AND PREDICTED JOINT TORQUES IN THE TRAJECTORY TEST

| method | Joint 1 | Joint 2 | Joint 3 | Joint 4 | Joint 5 | Joint 6 |
|---|---|---|---|---|---|---|
| CAD (N·m) | 0.0709 | 0.1342 | 0.1096 | 0.1138 | 0.0461 | 0.0334 |
| Fontanelli *et al.* (N·m) | 0.0635 | 0.1805 | 0.1313 | 0.1208 | 0.0589 | 0.0174 |
| Our method (N·m) | **0.0546** | **0.0908** | **0.0440** | **0.0910** | **0.0155** | **0.0172** |

### B. Gravity Compensation Validation

Two sets of experiments (Trajectory and Drift Tests) were conducted to validate the effectiveness of our gravity compensation method. These experiments involved extensive test configurations that cover the majority of the MTM workspace, which is defined as the collection of all possible configurations of the MTM within the joint ranges. We also conducted these tests with the CAD-based approach (CAD) by [6] and a method based on the work of Fontanelli *et al.* [11] and compared the results against ours. In the augmented method based on Fontanelli *et al.*, we reproduced their proposed method except that standard SLSE was applied for dynamic parameter estimation and identification data was the same with those for our method collected by the two-joint moving approach. Furthermore, we repeated these tests on eight MTMs in both JHU's and our groups to evaluate the robustness, reliability, and repeatability of our proposed method across different hardware.

*1) Trajectory Test:* In this experiment, the MTM moved along a trajectory going through ten randomly chosen joint configurations using a position controller. Fig. 8 shows the desired joint trajectory for this study. The MTM stayed at each desired configuration for 5 seconds before moving to the next one. All the joint states (such as position, velocity, and measured torque) were recorded throughout the process. GCC output torques were computed offline using the CAD, Fontanelli *et al.*, and our methods, and compared against the measured torques (see Fig. 9). The RMS relative error and maximum absolute error between the measured and predicted joint torques are computed for three approaches as shown in Tables II and III. During the computation, only the steady state output torques in each desired configuration were used.

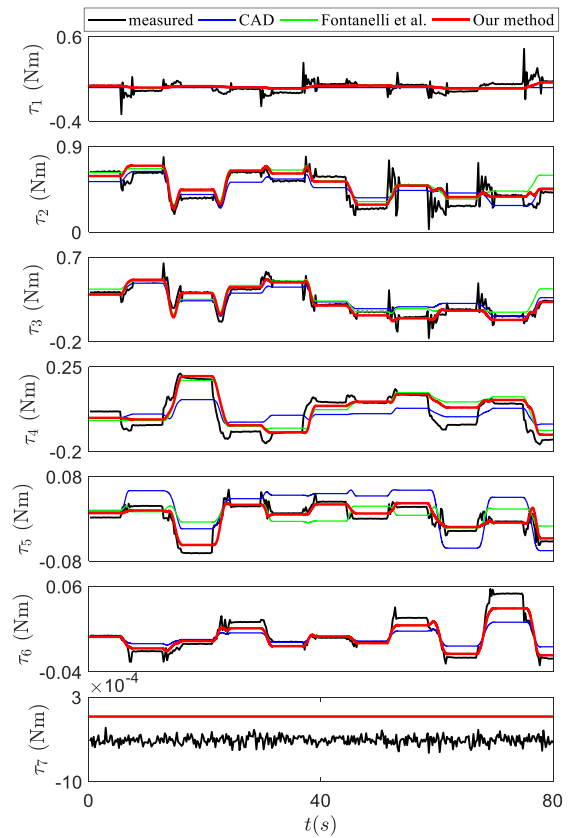

Fig. 9. Comparison of the GCC output joint torques obtained by CAD, Fontanelli *et al.* [11] and our method with measured torques.

As can be seen in Fig. 9, the GCC output torques generated through our method match very well with the measured joint torques, as compared to other methods (CAD, Fontanelli *et al.*). Moreover, Table II and Table III shows that these errors obtained through our approach for most joints are also lower than other approaches. However, it is still hard for us to judge if these error differences can create a catastrophic effect on the GCC unless we actually test the estimation results in the GCC experimentally. This echoes the need for the drift test over the entire workspace.

*2) Drift Test:* The purpose of the drift test is to quantify the performance of the GCC by measuring the drifting motion in joint space and task space. In this experiment, we first moved the MTM to a desired pose (position and orientation) using a position controller and then switched to the GCC for 2 seconds. Approximately 400 poses within the usable workspace of the MTM were randomly chosen for this study. All the joint states were recorded and used to compute the corresponding Cartesian position and orientation of the end-effector.

Fig. 10a shows the randomly generated sample points, which covered the majority of the workspace. Fig. 10b shows the drift results of one test sample, where both the translational and rotational drift errors of our method were much smaller than those of the other two methods. Fig. 10c and 10d show the normal distribution fit of both position and orientation drifts of the end-effector obtained through different methods, where the integrals of the normal distributions are equal to the number of sample points. It is notable that the means of the translational and rotational drift errors for our method (0.0046 m, 2.31°) were far smaller than those of the other methods (CAD: 0.0765 m, 73.16°; Fontanelli *et al.*: 0.0671 m, 43.01°) in the majority of workspace.

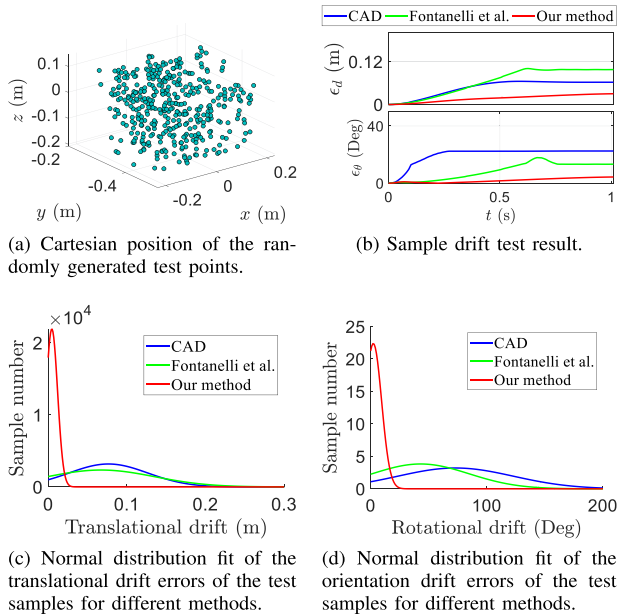

Fig. 10. Results of the drift test obtained by CAD, Fontanelli *et al.* [11] and our method.

## VII. CONCLUSION

We proposed a systematic approach for the gravity compensation of the dVRK MTM robotic arm with nonlinear external disturbance forces. This method used polynomial functions to model the nonlinear disturbance behavior, and integrated with the MLSE framework for parameter estimation and GCC. A two-joint moving data collection strategy was proposed and verified to be effective in reducing the estimation errors. Extensive experiments were conducted, and our proposed method demonstrated a significant improvement in precision, stability, and reliability compared to the previous approaches. The whole procedure was tested to be reproducible and reliable with eight different MTMs. A dVRK gravity compensation software package using this framework was published to the dVRK community. The developed framework can also apply to other serial manipulators with similar unmodeled dynamics. Future work will be the incorporation of the modeling of nonlinear hysteresis behavior of the electrical cables into our MLSE framework, with nonlinear optimization at each estimation iteration.

Although our work has achieved good gravity and external disturbance compensation performance, the hysteresis effect of cables (see Fig. 3) is not considered. Future work will be integrating hysteresis modeling into our MLSE framework.

ACKNOWLEDGMENT

The authors would like to thank all contributors to their library and dVRK software.